\documentclass{article}

\PassOptionsToPackage{numbers, compress}{natbib}

\usepackage[preprint]{neurips_2023}

\usepackage{longtable}
\usepackage[utf8]{inputenc} 
\usepackage[T1]{fontenc}   
\usepackage{hyperref}      
\usepackage{url}            
\usepackage{booktabs}       
\usepackage{amsfonts}       
\usepackage{CJKutf8}
\usepackage{nicefrac}       
\usepackage{microtype}      
\usepackage{xcolor}        
\usepackage{amsmath}
\usepackage{tcolorbox}
\usepackage{makecell} 
\usepackage{array}    
\usepackage{float}
\usepackage{graphicx}
\usepackage{ltablex}
\usepackage{lipsum}
\usepackage{graphicx}
\usepackage{wrapfig}
\usepackage[size=small]{caption}
\usepackage{mathtools}
\usepackage{tabularx}
\usepackage{amssymb}
\usepackage{amsthm}
\usepackage{soul}
\usepackage{float}

\definecolor{textblue}{rgb}{.2,.2,.7}
\definecolor{textred}{rgb}{0.54,0,0}
\definecolor{textblack}{rgb}{0,0,0}
\definecolor{textgreen}{rgb}{0,0.53,0}
\usepackage{listings}
\lstdefinestyle{pythonstyle}{
    language=Python,
    morekeywords={None}, 
    breaklines=true,
}

\usepackage{array}
\newcolumntype{M}[1]{>{\centering\arraybackslash}m{#1}}

\title{Octopus v4: Graph of language models}

\author{Wei Chen \thanks{Corresponding author} \\
Nexa AI\\
\texttt{\{alexchen\}@nexa4ai.com} \\
\And
Zhiyuan Li \\
Nexa AI\\
\texttt{\{zack\}@nexa4ai.com} \\
}

\begin{document}
\begin{CJK*}{UTF8}{gbsn}
\nocite{*}
\maketitle

\begin{abstract}
Language models have proven effective in a wide range of applications, yet the most sophisticated models are often proprietary. For example, GPT-4 by OpenAI and various models by Anthropic are expensive and consume substantial energy. In contrast, the open-source community has produced competitive models such as Llama3. Furthermore, smaller, niche-specific language models, such as those tailored for legal, medical, or financial tasks, have outperformed their proprietary counterparts. This paper introduces a novel approach that employs \textit{functional tokens} to integrate \textbf{multiple open-source models}, each optimized for a particular task. Our newly developed Octopus v4 model leverages \textit{functional tokens} to intelligently direct user queries to the most appropriate vertical model and reformat the queries to achieve the best performance. Octopus v4, an evolution of the Octopus v1, v2, and v3 models, excels in selection as well as in parameter understanding and reformatting. In addition, we explore the use of the graph as a versatile data structure that effectively coordinates multiple open-source models by harnessing the capabilities of the Octopus model and \textit{functional tokens}. Visit our open-source GitHub repository (\url{https://github.com/NexaAI/octopus-v4}) to try the Octopus v4 models (\url{https://huggingface.co/NexaAIDev/Octopus-v4}) and to contribute to a larger graph of language models. By activating models of around 10B parameters, we achieved a SOTA MMLU score of 74.8 among models at the same level.
\end{abstract}

\begin{figure}[h]
    \centering
    \includegraphics[width=0.8\textwidth]{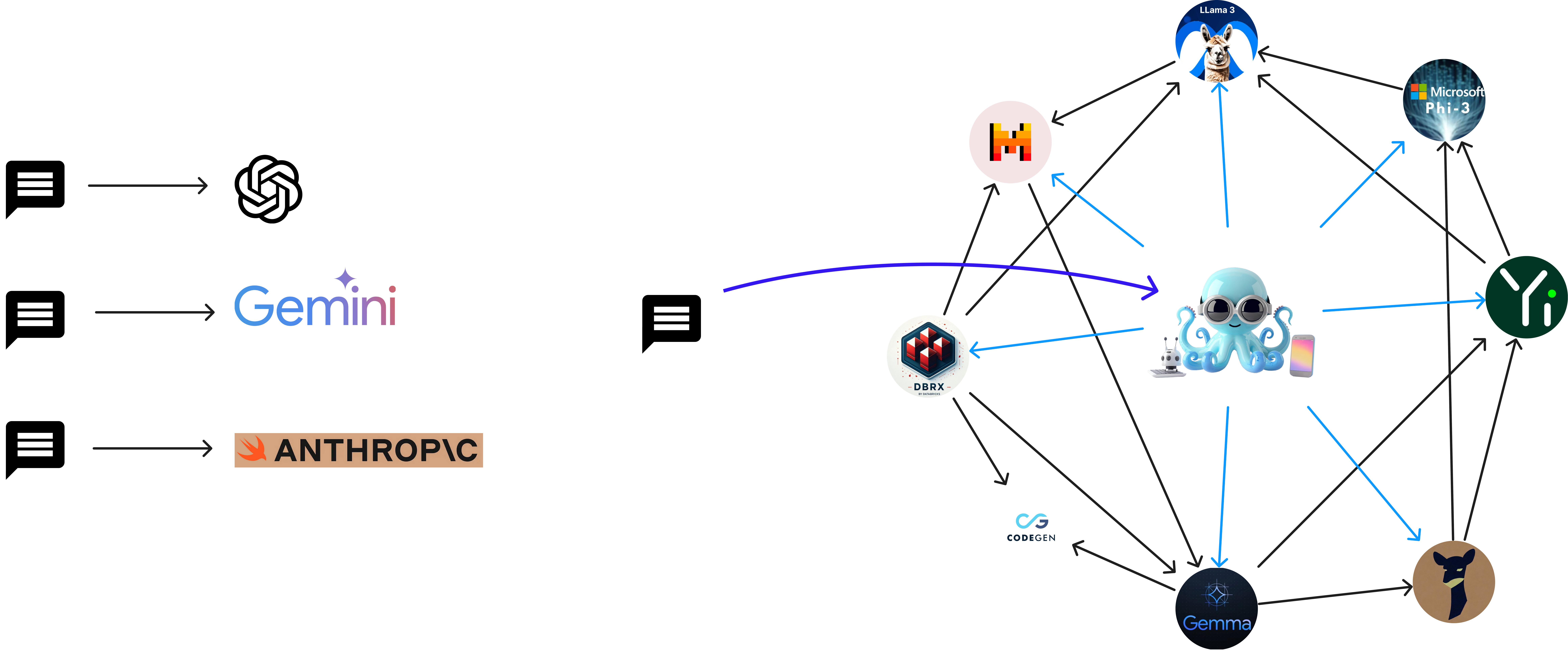}
    \caption{The shift from single-model inference, which employs a trillion-parameter model, to multi-node collaboration coordinated by the Octopus model. This framework optimizes the inference process by selecting the most suitable specialized models based on the user's query, activating only two models, each with fewer than 10B parameters, for one-step inference. Only a small graph is shown here, but the framework can support a large graph. See the demonstration of the graph here (\url{https://graph.nexa4ai.com/}).}
    \label{fig:transition}
\end{figure}

\section{Introduction}
The rapid progress in large language models (LLMs) has revolutionized natural language processing, enabling AI systems to understand and generate human-like text with remarkable accuracy. LLMs such as GPT-4 \cite{openai2024gpt4} and Anthropic's Claude \cite{claude32024}, trained on vast datasets, can capture the nuances of language, context, and meaning, leading to a wide range of potential use cases across industries. These models excel at various downstream tasks, such as highly accurate language translation, sentiment analysis to gauge customer opinions and monitor brand reputation, question answering drawing on extensive knowledge bases, and summarization of lengthy documents and articles. In healthcare, LLMs can process patient data and medical literature to support disease diagnosis, treatment planning, and drug discovery, even generating personalized patient reports \cite{sallam2023chatgpt, cascella2023evaluating}. The finance industry leverages LLMs for risk assessment, fraud detection, and automating financial report generation \cite{wu2023bloomberggpt, huang2023finbert}, while legal domains benefit from streamlined contract analysis, legal research, and document drafting \cite{cui2023chatlaw, kuppa2023chain}. LLMs also have significant potential in education \cite{sallam2023chatgpt, tang2024science}, providing personalized learning experiences, generating content, and offering instant feedback. As LLMs continue to advance, their impact across various domains is expected to grow significantly, automating tasks, enhancing decision-making, and driving innovation.

Since Meta's release of the Llama model and its successor, Llama 2 \cite{metallama2024}, in 2023, the open-source landscape for large language models (LLMs) has seen significant growth. This shift catalyzed the development of numerous innovative LLMs, released at an unprecedented rate. As key players in this dynamic field, these models have significantly influenced natural language processing. We highlight the most impactful open-source LLMs, including Mistral's sparse Mixture of Experts model Mixtral-8x7B \cite{mixtralexperts2024, jiang2023mistral}, Alibaba Cloud's multilingual Qwen1.5 series \cite{qwen}, Abacus AI's Smaug \cite{pal2024smaug}, and 01.AI's Yi models \cite{young2024yi}, which focus on data quality. Other notable models include Databricks' fine-grained MoE model DBRX \cite{dbrx2024}, Upstage AI's depth-upscaled SOLAR-10.7B \cite{kim2023solar}, Allen Institute for AI's alignment-focused TULU 2 and the collaborative OLMo series \cite{groeneveld2024olmo}, Microsoft's WizardLM powered by Evol-Instruct \cite{xu2023wizardlm}, Berkeley's Starling-7B \cite{starling2023}, Google DeepMind's Gemini-inspired Gemma models \cite{gemma-2023-open-models}, xAI's Grok \cite{grok12024}, and Deci AI's high-efficiency DeciLM-7B \cite{DeciFoundationModels}. In April 2024, we witnessed the most powerful open-source model to date, Meta's Llama3 \cite{llama3groq2024}, whose 70B-parameter version achieved an impressive inference speed of approximately 300 tokens per second using Groq \cite{llama3groq2024}. Shortly thereafter, more powerful open-source on-device models were released, including Microsoft's Phi-3-mini with 3.8 billion parameters \cite{abdin2024phi} and Apple's OpenELM family of models \cite{openelm}, which range from 1 to 3 billion parameters. These diverse models cater to various use cases, allowing users to select the optimal model for their needs.

Graph data structures have emerged as a powerful tool for representing complex relationships and dependencies in various domains. In computer science, a graph consists of a set of nodes (or vertices) connected by edges, which can be directed or undirected. This flexible structure allows for the representation of intricate connections and hierarchies that are difficult to capture using linear or tabular formats. Graphs offer several advantages over other data structures, including efficient traversal, pattern discovery, and the ability to model real-world networks. Many prominent companies have leveraged graph-based approaches to enhance their products and services. For example, Pinterest uses a graph structure to represent the relationships between users, boards, and pins, enabling personalized content recommendations and improved user engagement. Similarly, social networks like Facebook and LinkedIn rely on graph representations to model user connections, facilitating features such as friend suggestions and professional networking. In the context of integrating open-source language models, graph structures can be employed to represent the relationships between different models, their capabilities, and their optimal use cases. By treating each language model as a node in the graph and establishing edges based on their compatibility, complementary features, or task-specific performance, we can create a powerful framework for seamless model integration, intelligent query routing, and optimized performance.

The advent of on-device AI models has revolutionized the landscape of natural language processing, offering a host of advantages over traditional cloud-based approaches. These models, exemplified by Google's Gemma2B and the Llama7B model, are designed to run directly on user devices, ensuring data privacy by processing queries locally and eliminating the need for network communication with distant servers. This local processing not only enhances security but also reduces latency, enabling real-time interactions and improved user experiences. On-device AI agents, such as Octopus v2 \cite{chen2024octopus2, chen2024octopus} and v3 \cite{chen2024octopus3}, leverage these capabilities to provide intelligent assistance. However, the true potential of on-device AI lies in its seamless integration with cloud-based models, giving rise to the concept of \textbf{cloud-on-device collaboration} \cite{xu2018deeptype, sun2021mind}. By harnessing the power of both on-device and cloud-based models, AI systems can achieve unprecedented levels of performance, scalability, and flexibility. This collaboration allows for the efficient allocation of computational resources, with on-device models handling lighter and private tasks and cloud-based models tackling more complex or resource-intensive operations. Moreover, the Internet of Things (IoT) plays a crucial role in enabling this collaboration, connecting a vast network of devices and facilitating the exchange of data and insights between on-device and cloud-based models. The integration of on-device AI, cloud-based models, and IoT represents a paradigm shift in AI development. This approach combines the strengths of each component, creating a synergistic ecosystem that can adapt to the diverse needs of users and applications. As we continue to explore the potential of cloud-on-device collaboration, we can expect to see significant advancements in the field of AI.

In this paper, we introduce a new framework for using language models by constructing a graph with different vertical language models as its nodes. We leverage the capabilities of the Octopus v2 model and employ it as the coordinator. The transition from single-model inference to multi-node inference is illustrated in Figure (\ref{fig:transition}).

\section{Related work}

\textbf{Graph data format}\quad Graph algorithms have been a fundamental area of research in computer science, with a wide range of applications spanning from social network analysis to recommendation systems and bioinformatics. Classic graph algorithms, such as breadth-first search (BFS) and depth-first search (DFS), have been extensively studied and optimized for various graph representations. Dijkstra's shortest path algorithm and its variations have been crucial in solving routing problems and finding optimal paths in weighted graphs. The PageRank algorithm \cite{brin1998anatomy}, famously used by Google to rank web pages, has revolutionized the field of information retrieval and inspired numerous graph-based ranking techniques. Recent advancements in graph neural networks (GNNs) \cite{zhou2020graph, wu2020comprehensive, scarselli2008graph, xu2018powerful} have pushed the boundaries of graph-based learning, enabling the processing of graph-structured data for tasks such as node classification, link prediction, and graph generation. Frontier research in this area includes the development of more expressive and efficient GNN architectures, such as Graph Attention Networks (GATs) \cite{velickovic2017graph, brody2021attentive} and Graph Convolutional Networks (GCNs) \cite{zhang2019graph,wu2019simplifying}, which have achieved state-of-the-art performance on various graph-related tasks.

\textbf{Enhancing AI agents with functional tokens}\quad Building on the Octopus series (v1 to v3) \cite{chen2024octopus, chen2024octopus2, chen2024octopus3}, this research extends the capabilities of AI agents by utilizing functional tokens and uniting open-source models. These earlier versions effectively harnessed such tokens for advanced functionalities. We now investigate their extended use in integrating diverse open-source language models. Our studies indicate that functional tokens go beyond mere precision in classification tasks, such as selecting suitable functions or models for processing queries. Importantly, they amplify the Octopus model's ability to interpret and reshape user queries into an optimal format for the designated function, enhancing performance. This synergy between functional tokens and the Octopus models' capabilities in classification and query reformulation has been further applied to graph structures. Here, a pivotal aspect involves transferring information between nodes and selecting the appropriate neighbor for this transfer. Our enhanced Octopus model efficiently selects the best neighbor, restructures the information at the current node, and transmits optimized information to subsequent nodes.

\textbf{Multi-agent LLMs}\quad Multi-agent LLMs mark a pivotal evolution in AI, facilitating collaborative problem-solving through the integration of multiple specialized agents \cite{zhou2023multi}. Unlike traditional single-agent LLMs, these multi-agent systems harness collective intelligence from agents specialized in various domains. This collaborative approach yields more comprehensive solutions to complex issues. Multi-agent LLMs excel in delivering domain-specific expertise, enhancing problem-solving abilities, and offering robustness, reliability, and adaptability. These systems promise transformative impacts across sectors like healthcare, finance, education, and customer service by providing tailored expertise, personalized interactions, and efficient decision-making processes. However, deploying multi-agent LLMs involves challenges such as integration difficulties, data sharing issues, and maintaining smooth coordination between agents. Ongoing research into multi-agent LLMs is exploring possibilities like cross-domain expertise and real-time collaboration while considering ethical aspects. The adoption of graph architectures in our paper is also inspired by multi-agent systems. Advanced functionalities such as parallel function calling can be achieved through \textbf{self-connections}, and sequential action processing via \textbf{graph traversal}, enhancing operational efficiency and scalability.

\textbf{LLM scaling law}\quad Scaling laws \cite{kaplan2020scaling} for large language models (LLMs) have revolutionized our understanding of the relationship between model size, dataset size, computational resources, and performance. These laws indicate that larger models trained on vast datasets with ample computational power generally outperform smaller ones. However, as LLMs continue to scale up, they face significant challenges related to server capacity and power consumption, which limit their further expansion. Our proposed architecture addresses these scalability issues by leveraging distributed computing and node expansion techniques, enabling \textbf{nearly unlimited node scalability}. We can create a more powerful language model system by adding more nodes, bypassing the limitations imposed by server quantity and power supply.

\section{Methodology}

This section outlines the primary methods for incorporating language models as nodes within a graph and provides details on the system architecture tailored for real applications. It also discusses the training strategy for the Octopus model using a synthetic dataset. Finally, we highlight the system design for such a graph of language models in a production environment.

\subsection{Language model for classification from Octopus v2}
In the Octopus v2 paper, we introduced a method called the \textit{functional token} for classification within a fixed pool. The Octopus v2 model effectively handles the task of
\begin{equation}\label{equ:octopus}
P(f, params|q),
\end{equation}
where $f$ denotes a choice from the set $F$, and $params$ represents the reformulated information derived from the query $q$. The paper illustrates this method's application to the task of function calling. Additionally, the functional token can be adapted to other similar scenarios that require selecting the optimal choice from a specified pool and reformulating the query to transfer information to subsequent nodes. In typical use cases involving a predefined graph, each node, represented as a language model, has a fixed number of neighbors. To perform language model inference, the best neighboring node is selected, and information from the current node is passed to the next. Thus, the Octopus v2 model is well suited to handling this problem, demonstrating both rapid execution and high accuracy, as documented in the Octopus v2 paper.

\subsection{Language models as nodes in a graph}
\begin{figure}
    \centering
    \includegraphics[width=0.75\textwidth]{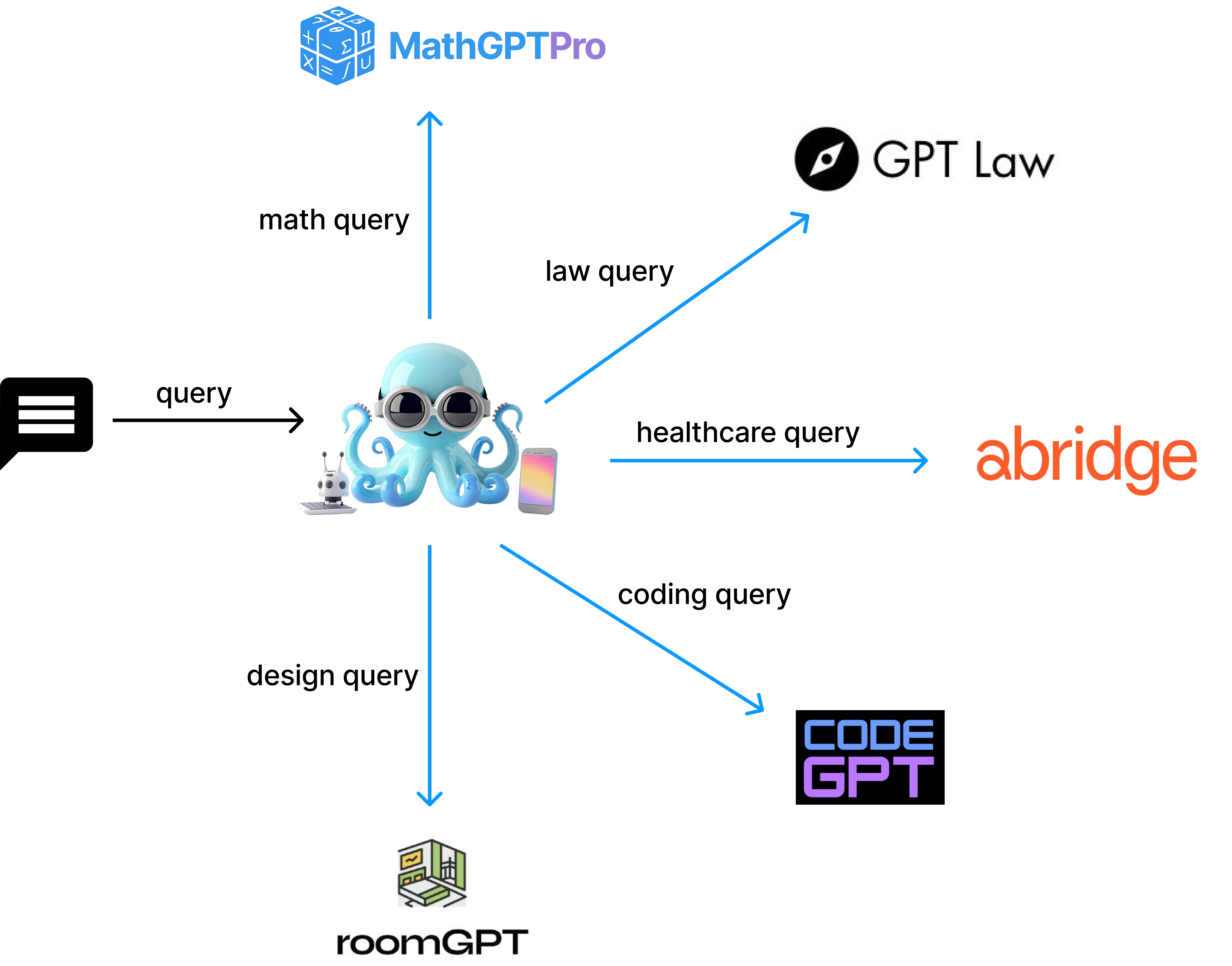}
    \caption{The Octopus model is utilized to determine the optimal neighboring node and generate appropriate information for transmission. Consider a scenario where the Octopus model's neighbors are MathGPT \cite{llama2mathgpt2024}, LawGPT \cite{cheng2024adapting}, HealthCareGPT \cite{abridge2024}, CodeGPT \cite{codegpt2024}, and RoomGPT \cite{roomsgpt2024}. The Octopus model can identify the most relevant GPT and transform the initial query into a format best suited for the selected GPT.}
    \label{fig:node_transfer}
\end{figure}

Consider a directed and heterogeneous graph defined as:
\begin{equation}
G = (N, E),
\end{equation}
where $N$ denotes the various nodes within the graph, and $E$ represents the edges that connect these nodes. Nodes are divided into two types: \textit{master nodes}, $N^m$, which coordinate queries by directing them to suitable \textit{worker nodes}, $N^w$, and transfer the information necessary for task execution; and worker nodes, which receive information from the master node and execute the required tasks, using an Octopus model to facilitate further coordination. The process of node information transfer is illustrated in Figure (\ref{fig:node_transfer}). For processing user queries $q$ and generating responses $y$, we model the probability as:
\begin{equation}
    P(y|q) = P(y|q; G).
\end{equation}

For a single-step task involving only one worker node, the procedure can be defined as follows:

\begin{equation}\label{equ:infer}
    P(y|q; G) = \underbrace{P(N^w, q_h|q; N^m)}_{\text{The Octopus v2 pattern problem}}P(y|q_h;N^w)
\end{equation}

Here, $P(N^w, q_h|q; N^m)$ uses an Octopus v2 model to select the best neighboring worker node for $N^m$ and to reformat the query into $q_h$, the reformulated query. This expression represents a typical problem that can be solved by the Octopus model, and it has the same structure as equation (\ref{equ:octopus}). The likelihood $P(y|q_h;N^w)$ is calculated by the language model situated at the worker node.

For multistep tasks, which are typical in multi-agent workflows, the process involves several sequential interactions among multiple nodes, as follows:
\begin{equation}
    P(y|q; G) = \prod_{i=0}^{k-1} \underbrace{P(N^w_i, q_{h_i}|q; N^m_i)}_{\text{The Octopus v2 pattern problem}}P(y|q_{h_i};N^w_i)
\end{equation}

This formula extends the single-step task to multiple steps, each handled by potentially different worker nodes and coordinated by their respective master nodes. Each step processes a part of the query and contributes to the final outcome, with $k$ representing the number of steps or interactions in the multi-agent process. This method exemplifies a coordination and execution pattern in a distributed AI system, leveraging the capabilities of multiple specialized agents within a graph-based framework.

The graph itself is predefined based on the available language models. Each worker model can also be an Octopus model capable of taking actions. For parallel function calling, the master node directs the query to the same node multiple times to execute the calls in parallel.

Compared to a large language model such as GPT-4, this design has the additional advantage that answering a single user query requires activating only two small language models rather than a trillion-parameter model. This translates into faster speed, lower energy cost, and fewer hardware demands. In Octopus v2, we demonstrated that functional tokens can replace RAG-based methods, achieving accurate function selection and fast generation. Thus, equation (\ref{equ:infer}) is executed quickly.

\subsection{Task planning using graphs for multistep operations}

\begin{figure}[H]
    \centering
    \includegraphics[width=0.75\textwidth]{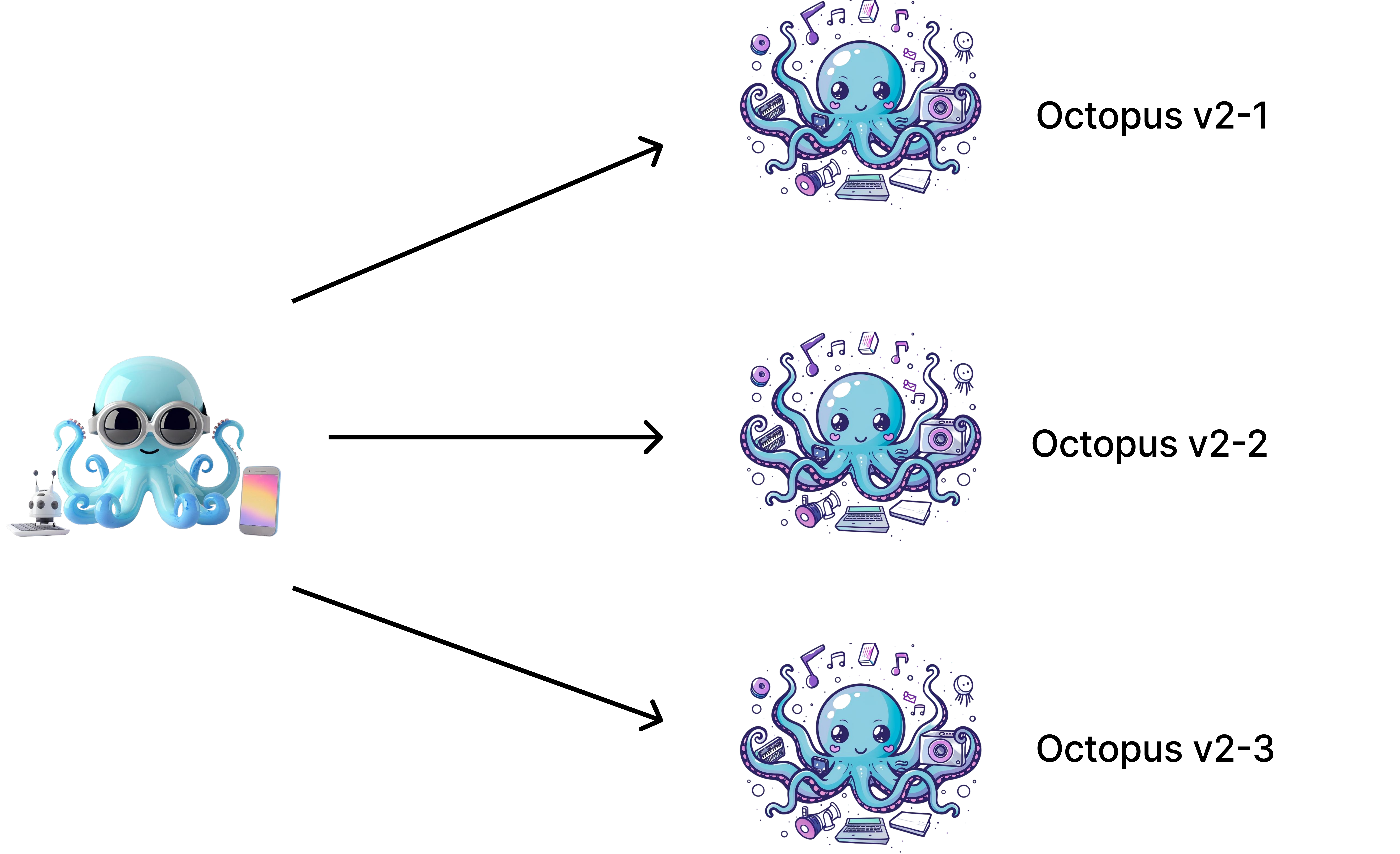}
    \caption{In our design, the architecture consists of two abstraction layers. The first layer utilizes functional tokens to represent the actions executable by the Octopus v2 model. This layer encompasses three distinct Octopus v2 models, each identified by different functional tokens, effectively differentiating them as separate AI agents. The second layer of abstraction pertains to the Octopus v4 model, where internal functional tokens are mapped to various v2 models. For simplicity, we include only three v2 models, but in real use cases one can map to many v2 models.}
    \label{fig:two_abstraction}
\end{figure}

In multistep task planning, incorporating numerous steps is essential. Traditionally, all available functions were listed in the context and submitted to a language model, which then generated plans based on user queries. This approach, however, faces limitations when the language model, especially one with fewer than 10B parameters, attempts to process lengthy function descriptions. Such models struggle to grasp extensive descriptions effectively. Moreover, this method does not consider the inherent relevance among different function descriptions. To address these challenges, constructing a graph that maps the relevance between various nodes (language models/agents) offers a viable solution. This graph-based approach ensures that only neighboring nodes relevant to a specific node are considered, significantly reducing the complexity of choices compared to the total number of function descriptions. By leveraging the capabilities of the Octopus v2 model, this strategy enhances efficiency, enabling rapid query redirection and reformatting. Our design in fact involves two layers of abstraction. First, for each language model, we can apply functional tokens to turn it into a single AI agent capable of performing individual function calls; alternatively, a single node/language model can be an ordinary language model, such as Llama3 or Phi3, that performs question answering, writing, and other tasks. The second layer of abstraction is that we can also create an Octopus v4 model to choose among the different nodes. The two layers of abstraction are illustrated in Figure (\ref{fig:two_abstraction}).

\subsection{Functional tokens and dataset collection}
As in the functional token architecture of Octopus v2, we conceptualize each model as a distinct function, utilizing functional tokens to activate specific model usage. This approach simplifies the function design for language models, requiring only a single input argument and a single output result. Additionally, for specific models, we can detail the required prompt template within the function's docstring. This allows the Octopus v4 model to restructure the original query to match the expected format. For instance, a function dedicated to processing legal information might be described as follows:

\begin{lstlisting}[style=pythonstyle]
def law_gpt(query):
    """
    A specialized language model equipped to handle queries related to legal studies, including international law, jurisprudence, and professional law. This model serves law students, practicing lawyers, and professionals in the legal field needing detailed legal explanations or interpretations. This model also reformats user queries into professional legal language.

    Parameters:
    - query (str): A detailed prompt that encapsulates a law-related question or issue. Speak in a professional legal manner.

    Returns:
    - str: Comprehensive legal analyses, solutions, or information related to the law query.
    """
\end{lstlisting}

Additionally, we construct the dataset using a strategy similar to that of the Octopus v2 paper. Following the methodology outlined there, we can train multiple functional tokens corresponding to various custom language models. As in the Octopus v2 paper, the dataset collection process relies on synthetic data to train the functional tokens. To better accommodate diverse queries, it may be beneficial to increase the temperature during data generation. This adjustment helps capture the variability and potential formatting inconsistencies of user queries, which are common in certain use cases.

\subsection{System design of language model graph}
This section details the system architecture of a complex graph in which each node represents a language model, utilizing multiple Octopus models for coordination. In preparation for production deployment, it is crucial to integrate a load balancer to manage system demands efficiently. Below, we divide the system into several manageable components, emphasizing the core methodologies:

\begin{itemize}
\item \textbf{Worker node deployment}:\quad Each worker node $N^w$ corresponds to an individual language model. We propose employing a serverless architecture for these nodes, specifically recommending Kubernetes \cite{kubernetes2019kubernetes} for its robust autoscaling capabilities based on memory usage and workload. We also limit the model parameters of all worker nodes to under 10B.
\item \textbf{Master node deployment}:\quad The master node should utilize a base model with fewer than 10B parameters (we use a 3B model in our experiments), enabling deployment on edge devices. Each worker node interfaces with an Octopus model for enhanced coordination. As demonstrated in Octopus v2, a compact LoRA model can be integrated to extend functional token capabilities. We suggest using a single base model supplemented by multiple LoRAs, one per worker node. The \href{https://github.com/predibase/lorax}{LoraX} library, an open-source initiative, is recommended for managing inference operations in this configuration.
\item \textbf{Communication}:\quad Worker and master nodes are distributed across various devices rather than confined to a single unit. Thus, internet connectivity is essential for transmitting data between nodes. While the master node may reside on a smart device, worker nodes are hosted in the cloud or on other devices, with results relayed back to the smart device. To support data caching needs, including chat history storage, we recommend \href{https://redis.io/glossary/distributed-caching/}{Redis} \cite{carlson2013redis}, a high-performance, in-memory database that facilitates distributed caching.
\end{itemize}

\begin{figure}
    \centering
    \includegraphics[width=1.0\textwidth]{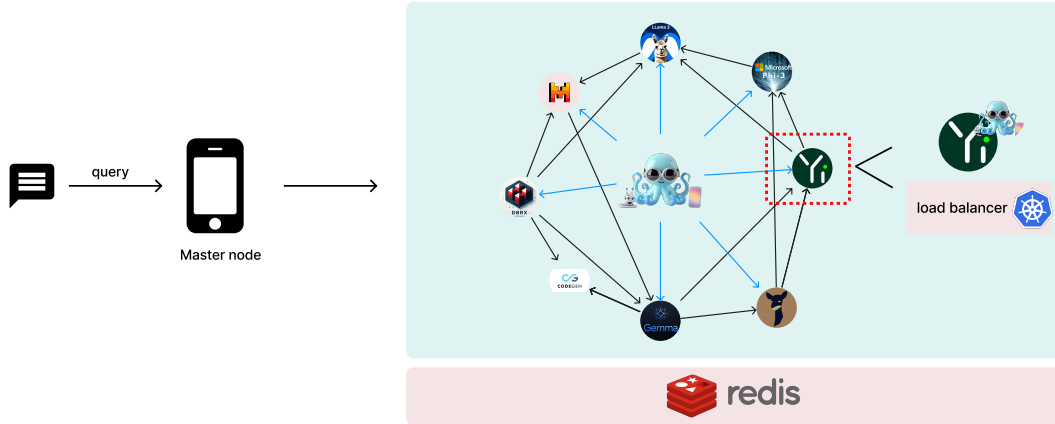}
    \caption{Our system design features a graph of language models with a master node deployed on a central device and worker nodes distributed across various devices. We employ Kubernetes (k8s) for serverless deployment of each individual worker language model. For efficient data sharing, we utilize a distributed caching mechanism supported by Redis. Note that each worker node has a small Octopus v4 LoRA attached to it to guide the selection of the next neighboring node in multi-agent use cases.}
    \label{fig:graph}
\end{figure}

The overall system design architecture is depicted in Figure (\ref{fig:graph}).

\section{Experiments}
In this section, we detail the experiments performed with our framework, which aim to enhance language model performance via multi-node collaboration. We demonstrate how our framework can improve language model efficacy using the MMLU benchmark \cite{hendrycks2020measuring} for evaluation. For this purpose, we apply 17 distinct models across the MMLU tasks. Upon receiving a user query, the Octopus v4 model directs the query to the relevant specialized model, which then reformats it suitably. The following experiment employs a simple graph; a more complex graph will be provided in our GitHub repository (\url{https://github.com/NexaAI/octopus-v4}) in the future, and building the ultimate graph will require the effort of the whole community.

\subsection{Specialized models}
The Massive Multitask Language Understanding (MMLU) benchmark encompasses 57 unique tasks, further categorized into 17 consolidated groups as recommended by the \href{https://github.com/hendrycks/test/blob/4450500f923c49f1fb1dd3d99108a0bd9717b660/categories.py}{authors}. Tasks such as graduate-level and high-school math have been grouped into the broader \textit{math} category. The tasks are divided as follows:
\begin{itemize}
\item \textbf{STEM}: Physics, Chemistry, Biology, Computer Science, Math, Engineering;
\item \textbf{Humanities}: History, Philosophy, Law;
\item \textbf{Social Sciences}: Politics, Culture, Economics, Geography, Psychology;
\item \textbf{Other}: Miscellaneous, Business, Health.
\end{itemize}

Specialized models were curated from Hugging Face based on benchmark scores, popularity, and user endorsements. Not all specialized tasks have corresponding models; for example, models for the Humanities and Social Sciences are notably absent. Nonetheless, the Llama3 model, adjusted with tailored system prompts, serves as a base model to simulate specialized capabilities without direct fine-tuning. The following 17 models are either specifically fine-tuned or customized through prompts:

\begin{itemize}
\item \textbf{Physics}: \textit{Weyaxi/Einstein-v6.1-Llama3-8B} (\url{https://huggingface.co/Weyaxi/Einstein-v6.1-Llama3-8B}), fine-tuned on a physics dataset (\url{https://huggingface.co/datasets/camel-ai/physics});
\item \textbf{Biology}: \textit{jondurbin/bagel-8b-v1.0} (\url{https://huggingface.co/jondurbin/bagel-8b-v1.0}), fine-tuned on a biology dataset;
\item \textbf{Computer Science}: \textit{Llama-3-Smaug-8B} (\url{https://huggingface.co/abacusai/Llama-3-Smaug-8B}), tailored for various computer science forums;
\item \textbf{Math}: \textit{Open-Orca/Mistral-7B-OpenOrca}, optimized for math (\url{https://huggingface.co/Open-Orca/Mistral-7B-OpenOrca});
\item \textbf{Engineering}: \textit{phi-2-electrical-engineering} (\url{https://huggingface.co/STEM-AI-mtl/phi-2-electrical-engineering}), fine-tuned on an electrical engineering dataset, selected for its relevance to MMLU;
\item \textbf{Law}: \textit{AdaptLLM/law-chat} (\url{https://huggingface.co/AdaptLLM/law-chat}), fine-tuned on a law dataset;
\item \textbf{Health}: \textit{AdaptLLM/medicine-chat} (\url{https://huggingface.co/AdaptLLM/medicine-chat}), optimized for medical data;
\item \textbf{Psychology, History, Philosophy, Politics, Culture, Geography, Business, Chemistry, Economics}: Currently, no specialized models are available for these areas. Custom system prompts and chain-of-thought (CoT) techniques are used with Llama3 to simulate specialized models;
\item \textbf{Other}: For the remaining tasks, the Phi3 model (\url{https://huggingface.co/microsoft/Phi-3-mini-128k-instruct}) is employed as a general-purpose model.
\end{itemize}

\subsection{MMLU benchmark evaluation}
This section presents a benchmark evaluation of the Octopus v4 system, comparing its performance with that of other renowned models on the MMLU benchmark to demonstrate our model's effectiveness. In our inference system, we utilize two compact language models: the 3B-parameter Octopus v4 model and a worker language model with no more than 8B parameters. The comparison is shown in Table~\ref{tab:mmlu}, and the inference procedure is illustrated in Figure (\ref{fig:node_transfer}).

An example user query is highlighted below:
\begin{tcolorbox}
Query: Tell me the result of derivative of $x^3$ when $x$ is 2?\\

Response: <nexa\_4> ('Determine the derivative of the function $f(x) = x^3$ at the point where $x$ equals 2, and interpret the result within the context of rate of change and tangent slope.')<nexa\_end>
\end{tcolorbox}
Note that \texttt{<nexa\_4>} is a functional token that maps to the math GPT.

\begin{table}[ht]
\centering
\caption{Comparison of MMLU (5-shot) scores between Octopus v4 and other models. During Octopus v4's inference, only two small language models, each with fewer than 10B parameters, are activated. Octopus v4 achieves a significant improvement in MMLU score at the cost of only a small number of additional tokens, owing to the use of functional tokens.}
\label{tab:mmlu}
\vspace{4pt}
\setlength{\tabcolsep}{14pt}
\begin{tabular}{@{}lc@{}}
\toprule
Model & MMLU, 5-shot (\%) \\
\midrule
Llama-3-8B-Instruct      & 68.4 \\
Phi-3-mini-128k-instruct & 68.1 \\
GPT-3.5                  & 70.0 \\
GPT-4                    & 86.4 \\
\midrule
Octopus v4               & 74.8 \\
\bottomrule
\end{tabular}
\end{table}

\section{Discussion and future work}
In this section, we present a brief tutorial on training a specialized model and outline our team's future plans.

\subsection{How to train a vertical model}
To effectively fine-tune a large language model for domain-specific expertise, begin by gathering a substantial corpus of high-quality, domain-relevant data. This collection should include textbooks, research papers, articles, and other pertinent materials that thoroughly cover the domain. It is crucial to ensure that the data is diverse, well organized, and embodies the domain knowledge intended for the model. Proceed by preprocessing this data—cleaning it, formatting it consistently, and addressing any specialized jargon or terminology.

Select a pre-trained large language model that suits your needs, and use the preprocessed domain-specific data for fine-tuning. This process adjusts the model's parameters to specialize in the chosen domain, effectively embedding the necessary expertise. Optionally, consider employing knowledge distillation to transfer insights from a larger model's API to a smaller, more efficient model. For this fine-tuning phase, the SFT trainer (\url{https://huggingface.co/docs/trl/sft_trainer}) provided by Hugging Face offers a user-friendly interface. We recommend using supervised fine-tuning followed by direct preference optimization.

\subsection{Future work}
Our current GitHub project focuses on developing a graph framework for language models and is in its initial phase. We plan to enhance this framework by integrating a variety of vertical-specific models and incorporating the advanced Octopus v4 models with multi-agent capability. Future releases will feature more robust graph representations in this repository, and the GitHub repository will be carefully maintained by Nexa AI. Unlike scaling laws, which are constrained by model and dataset size, our framework has no such limits, and we can create a large graph.

Additionally, we are developing Octopus 3.5, a multimodal model that processes vision, audio, and video data. Subsequent versions will incorporate this AI agent into our graph framework. Nexa AI also aims to develop compact, specialized models for diverse vertical domains.

\newpage
\medskip
{\small
\bibliographystyle{plainnat}
\bibliography{citation}
}


\section*{Appendix}
The functions used in our experiments are shown below:

\begin{lstlisting}[style=pythonstyle]
def physics_gpt(query):
    """
    A specialized language model designed to answer questions and provide insights on physics-related topics, including conceptual physics, college physics, high school physics, and astronomy. This model caters to learners at different educational stages, from high school to college levels. This model also reformats user queries into professional physics language.

    Parameters:
    - query (str): A detailed prompt that encapsulates a physics-related question or problem. It is designed to support a deep and professional discussion of physics topics.

    Returns:
    - str: Detailed explanations, solutions, or information related to the physics query.
    """


def chemistry_gpt(query):
    """
    A specialized language model tailored to assist with chemistry topics, including high school chemistry, college chemistry, and related chemical sciences. This tool aids students and researchers in deepening their understanding of chemical concepts and practices. This model also reformats user queries into professional chemistry language.

    Parameters:
    - query (str): A detailed prompt that encapsulates a chemistry-related question or problem. The language used is intended for a sophisticated exploration of chemistry.

    Returns:
    - str: Detailed explanations, solutions, or information related to the chemistry query.
    """


def biology_gpt(query):
    """
    This language model is dedicated to providing insights and answers on biology, encompassing high school biology, college biology, human anatomy, and related fields. It is an essential resource for students across educational levels and biology enthusiasts. This model also reformats user queries into professional biology language.

    Parameters:
    - query (str): A detailed prompt that encapsulates a biology-related question or problem, suitable for detailed and expert-level discussion.

    Returns:
    - str: Detailed explanations, solutions, or information related to the biology query.
    """


def computer_science_gpt(query):
    """
    Designed for computer science queries, this language model covers topics such as college computer science, high school computer science, computer security, and machine learning. It supports both academic and professional needs, enhancing learning and research in the field of computer science. This model also reformats user queries into professional computer science language.

    Parameters:
    - query (str): A detailed prompt related to computer science topics, suitable for academic and professional discussions.

    Returns:
    - str: Detailed responses that enhance understanding and provide solutions in computer science.
    """


def math_gpt(query):
    """
    A specialized language model designed to answer questions and provide insights on math-related topics, including abstract algebra, elementary mathematics, high school mathematics, college mathematics, and high school statistics. This model supports learners at various educational levels from high school to college. This model also reformats user queries into professional math language.

    Parameters:
    - query (str): A detailed prompt that encapsulates a math-related question or problem. Speak in a professional mathematician manner.

    Returns:
    - str: Detailed explanations, solutions, or information related to the math query.
    """


def electrical_engineering_gpt(query):
    """
    This language model offers expert guidance on electrical engineering topics, designed to support students, educators, and professionals in the field. It addresses questions related to fundamental and advanced electrical engineering concepts. This model also reformats user queries into professional electrical engineering language.

    Parameters:
    - query (str): A detailed prompt that encapsulates an electrical engineering-related question or problem, fostering professional-level discussions.

    Returns:
    - str: Comprehensive responses, solutions, or information related to the electrical engineering query.
    """


def history_gpt(query):
    """
    A specialized language model designed to answer questions and provide insights on history-related topics. This model covers a broad range of historical subjects including high school European history, high school US history, high school world history, and prehistory. It aims to support learners and enthusiasts from various educational backgrounds. This model also reformats user queries into professional history language.

    Parameters:
    - query (str): A detailed prompt that encapsulates a history-related question or problem. Speak in a manner suited for historians or history students.

    Returns:
    - str: Detailed explanations, historical analyses, or information related to the history query.
    """


def philosophy_gpt(query):
    """
    A specialized language model designed to provide expert responses on various philosophy-related topics, including formal logic, logical fallacies, moral disputes, moral scenarios, and world religions. This model is useful for students, educators, and philosophy enthusiasts seeking deep philosophical discussions and insights. This model also reformats user queries into professional philosophy language.

    Parameters:
    - query (str): A detailed prompt that encapsulates a philosophy-related question or problem. Speak in a professional philosopher manner.

    Returns:
    - str: In-depth philosophical analysis or discussions relevant to the query.
    """


def law_gpt(query):
    """
    A specialized language model equipped to handle queries related to legal studies, including international law, jurisprudence, and professional law. This model serves law students, practicing lawyers, and professionals in the legal field needing detailed legal explanations or interpretations. This model also reformats user queries into professional legal language.

    Parameters:
    - query (str): A detailed prompt that encapsulates a law-related question or issue. Speak in a professional legal manner.

    Returns:
    - str: Comprehensive legal analyses, solutions, or information related to the law query.
    """


def politics_gpt(query):
    """
    A specialized language model designed to delve into topics related to politics and public relations, including high school government and politics, security studies, and US foreign policy. This model aids political science students, professionals, and enthusiasts in gaining a better understanding of political dynamics and theories. This model also reformats user queries into professional politics language.

    Parameters:
    - query (str): A detailed prompt that encapsulates a politics-related question or discussion. Speak in a manner suitable for political analysts.

    Returns:
    - str: Detailed political analysis, insights, or information pertaining to the politics query.
    """


def culture_gpt(query):
    """
    A specialized language model designed to explore cultural and societal topics, particularly focusing on human sexuality and sociology. This model is ideal for cultural studies students, sociologists, and anyone interested in understanding the dynamics of human societies and cultures. This model also reformats user queries into professional sociocultural analyst language.

    Parameters:
    - query (str): A detailed prompt that encapsulates a culture-related question or topic. Speak in a professional sociocultural analyst manner.

    Returns:
    - str: Detailed cultural insights, analyses, or information related to the cultural query.
    """


def economics_gpt(query):
    """
    A specialized language model designed to tackle questions and provide insights into economics, including econometrics, high school macroeconomics, and high school microeconomics. This model assists students, economists, and financial analysts in understanding economic theories and applications. This model also reformats user queries into professional economics language.

    Parameters:
    - query (str): A detailed prompt that encapsulates an economics-related question or problem. Speak in a manner suitable for economists.

    Returns:
    - str: Detailed economic explanations, analyses, or solutions relevant to the economics query.
    """


def geography_gpt(query):
    """
    A specialized language model developed to address inquiries related to geography, specifically focusing on high school geography. This model supports students and educators in understanding geographical concepts, theories, and real-world applications. This model also reformats user queries into professional geography language.

    Parameters:
    - query (str): A detailed prompt that encapsulates a geography-related question or topic. Speak in an educational manner suitable for geographers.

    Returns:
    - str: Detailed geographical information, analyses, or insights related to the geography query.
    """


def psychology_gpt(query):
    """
    A specialized language model focused on providing expert responses on topics related to psychology, including high school psychology, professional psychology, and human aging. This model is particularly valuable for psychology students, clinicians, and researchers seeking to understand various psychological theories and practices. This model also reformats user queries into professional psychologist language.

    Parameters:
    - query (str): A detailed prompt that encapsulates a psychology-related question or discussion. Speak in a professional psychologist manner.

    Returns:
    - str: In-depth psychological analyses, solutions, or information relevant to the psychology query.
    """


def business_gpt(query):
    """
    A specialized language model designed to address topics related to business, including business ethics, management, and marketing. This model supports business students, professionals, and entrepreneurs in understanding business practices, theories, and market dynamics. This model also reformats user queries into professional business language.

    Parameters:
    - query (str): A detailed prompt that encapsulates a business-related question or problem. Speak in a professional business manner.

    Returns:
    - str: Detailed business insights, strategies, or information relevant to the business query.
    """


def health_gpt(query):
    """
    A specialized language model designed to provide answers and insights on health-related topics, including anatomy, clinical knowledge, college medicine, medical genetics, nutrition, and virology. This model assists medical students, health professionals, and researchers in understanding complex medical and health issues. This model also reformats user queries into professional medical language.

    Parameters:
    - query (str): A detailed prompt that encapsulates a health-related question or issue. Speak in a professional medical manner.

    Returns:
    - str: Detailed medical explanations, solutions, or information related to the health query.
    """


def general_gpt(query):
    """
    A general-purpose language model designed to provide answers and insights across a wide array of topics not specifically categorized under other specialized models. This tool is specifically useful for users seeking information on miscellaneous and diverse topics that do not fall into the standard academic or professional categories such as physics, chemistry, biology, computer science, math, electrical engineering, history, philosophy, law, politics, culture, economics, geography, psychology, business, or health.

    Parameters:
    - query (str): A general prompt encompassing any topic of interest outside the specified categories. Speak in a broad and inclusive manner.

    Returns:
    - str: Comprehensive explanations or information pertaining to the general query, ensuring a focus away from the excluded fields.
    """
\end{lstlisting}

\end{CJK*}
\end{document}